\documentclass[letterpaper, 10 pt, conference]{ieeeconf}  

\IEEEoverridecommandlockouts                              

\usepackage{tabularx}
\usepackage{multirow}
\usepackage{tabularx}
\usepackage{optidef}
\usepackage{mathtools}
\usepackage{color}
\usepackage{xspace}

\usepackage{amsmath}
\usepackage{graphicx}
\newcommand*{\Scale}[2][4]{\scalebox{#1}{$#2$}}%
\providecommand{\FullStop}{\text{~\@.\xspace}}
\providecommand{\Comma}{\text{~,\xspace}}
\usepackage{bm}
\graphicspath{figure}

\title{\LARGE \bf
Hierarchical Motion Planning and Offline Robust Model Predictive Control for Autonomous Vehicles
}

\author{Hung Duy Nguyen$^{1}$, Minh Nhat Vu$^{1,2}$, Nguyen Ngoc Nam$^{3,4}$, and Kyoungseok Han$^{3}$
\thanks{$^{1}$Automation \& Control Institute (ACIN), TU Wien, Vienna, 1040, Austria
        ({\{nguyen, vu\}@acin.tuwien.ac.at})}%
\thanks{$^{2}$Austrian Institute of Technology GmbH (AIT), 1210, Vienna, Austria
        ({minh.vu@ait.ac.at})}        
\thanks{$^{3}$School of Mechanical Engineering, Kyungpook National University, Daegu, 41566, South Korea
        ({\{nnnam, kyoungsh\}@knu.ac.kr})}
\thanks{$^{4}$Faculty of Electrical and Electronic Engineering, Phenikaa University, Hanoi, 12116, Vietnam
        ({nam.nguyenngoc@phenikaa-uni.edu.vn})}
}

\begin{document}

\maketitle
\thispagestyle{empty}
\pagestyle{empty}

\begin{abstract}

Driving vehicles in complex scenarios under harsh conditions is the biggest challenge for autonomous vehicles (AVs).
To address this issue, we propose hierarchical motion planning and robust control strategy using the front active steering system in complex scenarios with various slippery road adhesion coefficients while considering vehicle uncertain parameters.
Behaviors of human vehicles (HVs) are considered and modeled in the form of a car-following model via the Intelligent Driver Model (IDM).
Then, in the upper layer, the motion planner first generates an optimal trajectory by using the artificial potential field (APF) algorithm to formulate any surrounding objects, e.g., road marks, boundaries, and static/dynamic obstacles.
To track the generated optimal trajectory, in the lower layer, an offline-constrained output feedback robust model predictive control (RMPC) is employed for the linear parameter varying (LPV) system by applying linear matrix inequality (LMI) optimization method that ensures the robustness against the model parameter uncertainties.
Furthermore, by augmenting the system model, our proposed approach, called offline RMPC, achieves outstanding efficiency compared to three existing RMPC approaches, e.g., offset-offline RMPC, online RMPC, and offline RMPC without an augmented model (offline RMPC w/o AM), in both improving computing time and reducing input vibrations.

\end{abstract}

\section{Introduction}
In recent years, driving autonomous vehicles (AVs) under adverse road surfaces, see, \cite{zang2019impact}, including rain, snow, fog, and hail, where the road adhesion coefficient is low, has been a massive challenge and barrier. Hence, an advanced control strategy is urgently required to achieve tracking performance and vehicle stability.

As known widely, the AV system is constructed primarily on four main functional modules, i.e., environment perception, decision-making, motion planning, and control algorithm, see, e.g., \cite{nguyen2023risk, liu2022interaction, nguyen2023safe}. Environment perception and motion control are considered the brain of the autonomous system, see, e.g., \cite{hang2020human, teng2023motion}. In contrast, motion planning and control are critical components of an autonomous system's ability, see, e.g., \cite{vu2022fast,nguyen2023risk}, to navigate and interact with its environment safely and effectively. Thus, these modules should be carefully designed based on the behavior of objects in the traffic environment.

Motion planning algorithms play an important role in navigating to avoid collisions and provide feasible trajectories for controllers. Tree-based path-planning algorithms, e.g., Dijkstra, RRT, RRT*, and A*, are proposed to generate the shortest path from the starting point to the goal without collisions, see, \cite{gonzalez2015review}. However, the computational burden has been an issue when applied to the automotive field. To address this issue, the artificial potential field (APF) algorithm, see, e.g., \cite{nguyen2023linear,vu2020fast,vu2022sampling}, is suggested to reduce the computational complexity while generating the short path by formulating the obstacle's potential values. In this manner, during driving, an optimal trajectory is generated when considering interactions of road objects, i.e., road marks, boundaries, and static/dynamic obstacles.

Additionally, the control level can be considered the last step of an autonomous system to follow the generated trajectory at the previous level. Model predictive control (MPC) has been employed recently as an advanced controller, see, e.g., \cite{nguyen2023risk, kim2023state,vu2022fast}, considering the input and output constraints. Then, by minimizing the objective function, the AV shows outstanding performance in tracking and stability when compared with conventional controllers, see, \cite{nguyen2023linear}.

Although these approaches handle vehicle tracking and stability problems well, uncertain parameters and complex scenarios are still massive challenges, see, \cite{teng2023motion}. Therefore, to address these challenges, our study proposes a hierarchical strategy, consisting of upper and lower layers. First, the upper layer deals with complex scenarios by generating an optimal trajectory via the APF algorithm that detects traffic infrastructure objects and static/dynamic obstacles. Further, linear matrix inequality (LMI) optimization-based robust model predictive control (RMPC) is employed at the lower layer to handle uncertain vehicle parameters. In this manner, two huge challenges of complex scenarios and uncertain parameters are solved well while ensuring the vehicle tracking performance and stability to avoid collisions when driving on a slippery road.

The two main contributions of this paper are outlined in the following: (i) An optimal trajectory is generated by formulating obstacle potential values via the APF, see, \cite{nguyen2023linear}; therefore, the AV can avoid different obstacles in arbitrary complex scenarios; (ii) Furthermore, by augmenting the vehicle model, we handled the steering wheel angle rate to improve the input's vibrations and so helped the AV improve the stability ability when driving on various road adhesion coefficients with a relatively high speed. In this manner, our proposed approach emphasized efficiency when compared with the offset-offline RMPC method, online RMPC method, and offline RMPC method without an augmented model (offline w/o AM), see, e.g., \cite{wan2003efficient, nam2023robust}.

\section{System Modeling}
\subsection{Traffic Environment Model}
The surrounding HV's driving behaviors are modeled by a car-following model using IDM, see, \cite{treiber2000congested}. The vehicle acceleration of each $i^{th}$ HV is calculated as
\begin{equation}
    \dot v_{\left( {{s^i},{v^i},\Delta {v^i}} \right)}^i = {a^i}\left[ {1 - {{\left( {\frac{{{v^i}}}{{v_o^i}}} \right)}^\delta } - {{\left( {\frac{{s_{\left( {{v^i},\Delta {v^i}} \right)}^{i*}}}{{{s^i}}}} \right)}^2}} \right],
\end{equation}
\noindent where $a$, $v_o$, and $\delta$ denote the maximum acceleration, desired speed, and free acceleration exponent, respectively; $s^i = x^{i-1} - x^i - l$ is the relative distance between the ${(i-1)}^{th}$ preceding car and the $i^{th}$ following car while $l$ denotes the length of car; and $\Delta v^i = v^i - v^{i-1}$ presents the relative longitudinal velocity. Besides, $s_{\left( {{v^i},\Delta {v^i}} \right)}^{*}$ denotes the desirable gap, which is calculated as follows:
\begin{equation}
    s_{\left( {{v^i},\Delta {v^i}} \right)}^{i*} = s_o^i + {v^i}{T_{{\text{gap}}}} + \frac{{{v^i}\Delta {v^i}}}{{2\sqrt {{a^i}{b^i}} }}\Comma
\end{equation}
\noindent where $b$ and $T_{\text{gap}}$ are the desirable deceleration and time gap.

\subsection{Path Tracking Model}
The error dynamics model for lateral trajectory and heading angle, see, \cite{rajamani2011vehicle}, is defined in the following:
\begin{subequations} \label{eq.error.dyn}
    \begin{align}
        & {{\dot e}_y} = \dot y - {{\dot y}_{\text{ref}}} = \dot y + v_x e_{\psi}\Comma \\
        & {{\dot e}_\psi } = \dot \psi  - {{\dot \psi }_{\text{ref}}}\FullStop \label{eq.psi_dd.error}
    \end{align}
\end{subequations}

By combining the state-space representation of the single-track model with \eqref{eq.error.dyn}, the linear vehicle tracking model system, see, \cite{rajamani2011vehicle}, is rewritten as
\begin{subequations} \label{eq.error.model}
    \begin{align}
        & \begin{array}{l}
{{\ddot e}_y} = \frac{{2\left( {{C_f} + {C_r}} \right)}}{m}{e_\psi } - \frac{{2\left( {{C_f} + {C_r}} \right)}}{{m{v_x}}}{{\dot e}_y} - \frac{{2\left( {{l_f}{C_f} - {l_r}{C_r}} \right)}}{{m{v_x}}}{{\dot e}_\psi }\\
\begin{array}{*{20}{c}}
{}&{}
\end{array} - \left( {\frac{{2\left( {{l_f}{C_f} - {l_r}{C_r}} \right)}}{{m{v_x}}} + {v_x}} \right){{{\dot \psi} }_{{\text{ref}}}} + \frac{{2{C_f}}}{m}{{\delta} _f}\Comma
\end{array} \\
        & \begin{array}{l}
{{\ddot e}_\psi } = \frac{{2\left( {{l_f}{C_f} - {l_r}{C_r}} \right)}}{{{I_z}}}{e_\psi } - \frac{{2\left( {{l_f}{C_f} - {l_r}{C_r}} \right)}}{{{I_z}{v_x}}}{{\dot e}_y} + \frac{{2{l_f}{C_f}}}{{{I_z}}}{{\delta} _f}\\
\begin{array}{*{20}{c}}
{}&{}
\end{array} - \frac{{2\left( {l_f^2{C_f} + l_r^2{C_r}} \right)}}{{{I_z}{v_x}}}{{{\dot \psi} }_{{\text{ref}}}} - \frac{{2\left( {l_f^2{C_f} + l_r^2{C_r}} \right)}}{{{I_z}{v_x}}}{{\dot e}_\psi }\Comma
\end{array}
    \end{align}
\end{subequations}
\noindent where $m$, $I_z$, $v_x$, and $\dot e_{\psi}$ represent the vehicle total mass, the vehicle inertia moment, longitudinal velocity, and yaw rate error; $l_f$ and $l_r$ denote distances between the front and rear axles to the vehicle's center of gravity; $C_f$ and $C_r$ represent the front/rear tire cornering stiffness, respectively.

We present a state-space equation of the vehicle tracking error model \eqref{eq.error.model} in discrete time, defined as follows:
\begin{equation} \label{eq.discrete.veh.model}
{\bm{\xi}}_{\text{error}} (t+1) = \mathbf{A}_{\text{d}} {\bm{\xi}}_{\text{error}} (t) + \mathbf{B}_{\text{d}} {u}(t) + \mathbf{E}_{\text{d}} {{\dot \psi} _{\text{ref}}}\Comma
\end{equation}
\noindent where ${\bm{\xi}}_{\text{error}} = [e_y, {\dot e}_y, e_{\psi}, {\dot e}_{\psi}]^{{\bm{\top}}}$ is the state variables; ${u} = {\delta} _f$ denotes the input control signal. Additionally, the discrete system matrices (i.e., $\mathbf{A}_{\text{d}}$, $\mathbf{B}_{\text{d}}$, and $\mathbf{E}_{\text{d}}$) can be found in \cite{rajamani2011vehicle}.

Let us define $\Delta {u}(t) = {u}(t) - {u}(t-1)$, the discrete-time model \eqref{eq.discrete.veh.model} is transferred into the following extended model:
\begin{equation} \label{eq.ext.discrete.veh.model}
{\bm{\xi}_\text{ext}} (t+1) = \mathbf{A}_\text{ext} {\bm{\xi}_\text{ext}} (t) + \mathbf{B}_\text{ext} {\Delta} {u}(t) + \mathbf{E}_\text{ext} {{\dot \psi} _{\text{ref}}}\Comma
\end{equation}
\noindent where ${\bm{\xi}_\text{ext}} (t) = {\left[ {{{\bm{\xi}} _{\text{error}}}(t), {u}(t)} \right]^{{\bm{\top}}}}$ denotes the extended model state variables; ${\Delta} {u}$ now is the control input command. Additionally, the extended model matrices are obtained as
\begin{equation}
    \Scale[0.95]{\mathbf{A}_\text{ext} = \left[ {\begin{array}{*{20}{c}}
    {{\mathbf{A}_{\text{d}}}}&{{\mathbf{B}_{\text{d}}}}\\
    \mathbf{0}&\mathbf{I}
    \end{array}} \right] , \mathbf{B}_\text{ext} = \left[ {\begin{array}{*{20}{c}}
    \mathbf{B}_{\text{d}}\\
    \mathbf{I}
    \end{array}} \right] , \mathbf{E}_\text{ext} = \left[ {\begin{array}{*{20}{c}}
    {\mathbf{E}_{\text{d}}}\\
    \mathbf{0}
    \end{array}} \right] \FullStop }
\end{equation}

Let us augment the model as $\bm{\xi} \left( {k} \right) = \left[ {\bm{\xi}_\text{ext} \left( {k} \right), {\dot \psi} _{\text{ref}}} \left( {k} \right) \right]^{\top}$ while assuming approximately the yaw rate reference as ${\dot \psi} _{\text{ref}} \left( {k+1} \right) = {\dot \psi} _{\text{ref}} \left( {k} \right)$, the dynamics system can be defined the augmented model as follows:
\begin{equation} \label{eq.augmented.system}
    \bm{\xi} \left( {t + 1} \right) = \underbrace {\left[ {\begin{array}{*{20}{c}}
    \mathbf{A}_\text{ext}&\mathbf{E}_\text{ext}\\
    \mathbf{0}&\mathbf{I}
    \end{array}} \right]}_{{\mathbf{A}}}\bm{\xi} \left( {t} \right) + \underbrace {\left[ {\begin{array}{*{20}{c}}
    \mathbf{B}_\text{ext}\\
    \mathbf{0}
    \end{array}} \right]}_{{\mathbf{B}}} {\Delta} {u}\left( {t} \right) \FullStop
\end{equation}

\subsection{Linear Parameter Varying System}
When driving under different pavement coefficients, the wheel is always in contact with the road surface, resulting in an uncertain tire stiffness coefficient. Therefore, we can assume that the uncertain tire stiffness coefficients at the front and rear wheels are in some specific boundaries as follows:
\begin{equation}
  {\bar C_{f/r}}/\kappa  \le {C_{f/r}} \le \kappa  \times {\bar C_{f/r}}\Comma
\end{equation}
\noindent where $\bar C_{f/r}$ denotes the nominal values. Besides, $\kappa$ is a tunable constant value that characterizes uncertain parameters.

Based on these uncertain parameters, we rewrite the augmented model \eqref{eq.augmented.system} under the LPV system, see, \cite{park2011output}, as
\begin{equation} \label{eq.dis.veh.model}
    {\bm{\xi}} \left( t + 1 \right) = \mathbf{A}\left( {\rho \left( t \right)} \right){\bm{\xi}} \left( t \right) + \mathbf{B}\left( {\rho \left( t \right)} \right) {\Delta} {u}\left( t \right),
\end{equation}
\noindent where $\rho(t)$ characterizes the uncertainty of parameter varying at each time step $t$. Therefore, the discretized uncertain matrices $\left[ {\mathbf{A}\left( {\rho \left( t \right)} \right),\mathbf{B}\left( {\rho \left( t \right)} \right)} \right]$ is assumed to be bounded and they belong to the polytopic set as $\left[ {\mathbf{A}\left( {\rho \left( t \right)} \right),\mathbf{B}\left( {\rho \left( t \right)} \right)} \right] \in \Omega$, where $\Omega  = {C_o}\left\{ {\left[ {{\mathbf{A}(1)},{\mathbf{B}(1)}} \right], \ldots ,\left[ {{\mathbf{A}(j)},{\mathbf{B}(j)}} \right]} \right\}$ denotes the convex hull, while ${\left[ {\mathbf{A}(j),\mathbf{B}(j)} \right]}$ represents vertices of the polytopic set when $j = 1, \dots , 4$ corresponding to the obtained matrices by considering maximum and minimum values of front and rear tire cornering stiffness.

\section{Problem Formulation And Hierarchical Framework}
\begin{figure}[t!]
    \centering
    \includegraphics[width = 85mm]{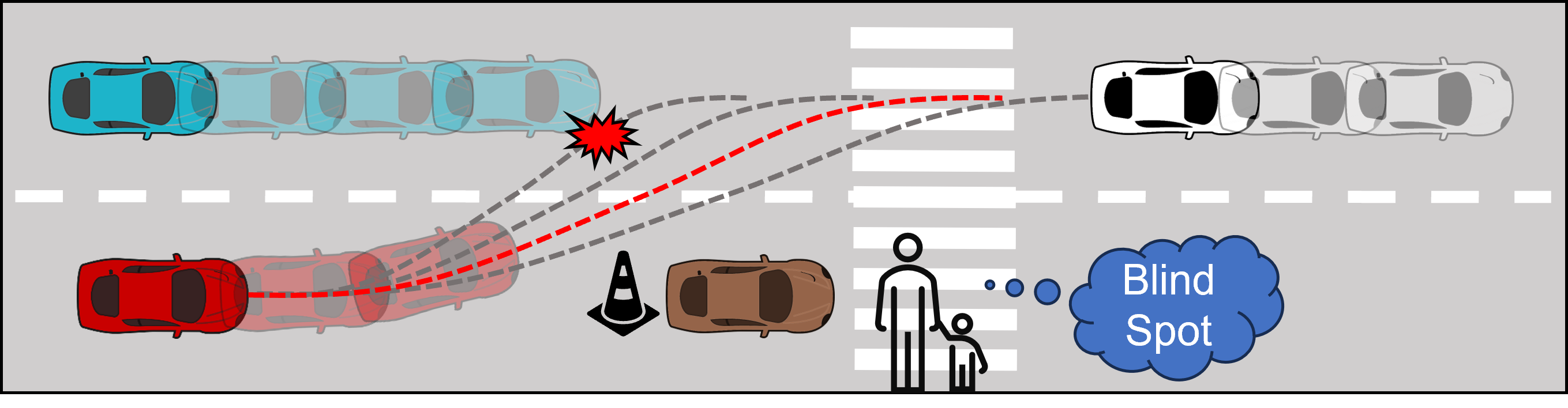}
    \caption{\label{fig.scenario} Schematic of highway driving strategy.}
\end{figure}

\subsection{Problem Formulation}
This study addresses one of the hardest traffic environments when the AV drives in complex scenarios under various road adhesion coefficients. More specifically, the AV aims to prevent car crashes on the road in emergencies, depending on each specific situation, by controlling vehicle steering to track the optimal trajectory.

A complex traffic maneuver is proposed in Fig.~\ref{fig.scenario} where surrounding objects are considered comprehensively, i.e., behaviors of HVs and pedestrians. The AV will perform the lane-change action as soon as the forward obstacle is observed and the front-end crash is expected. Additionally, during the lane-changing period, many risks may arise; in particular, two typical cases that can cause challenges are as follows: (i) While performing lane-changing action, another HV, located on the adjacent lane, drives at a relatively high speed, which leads to an aggressive scenario; (ii) In an unexpected case, a pedestrian, which is assumed to be in a blind spot where the AV cannot observe it, suddenly crosses the road leading to an unexpected scenario.

\begin{figure}[t!]
    \centering
    \includegraphics[width = 70mm]{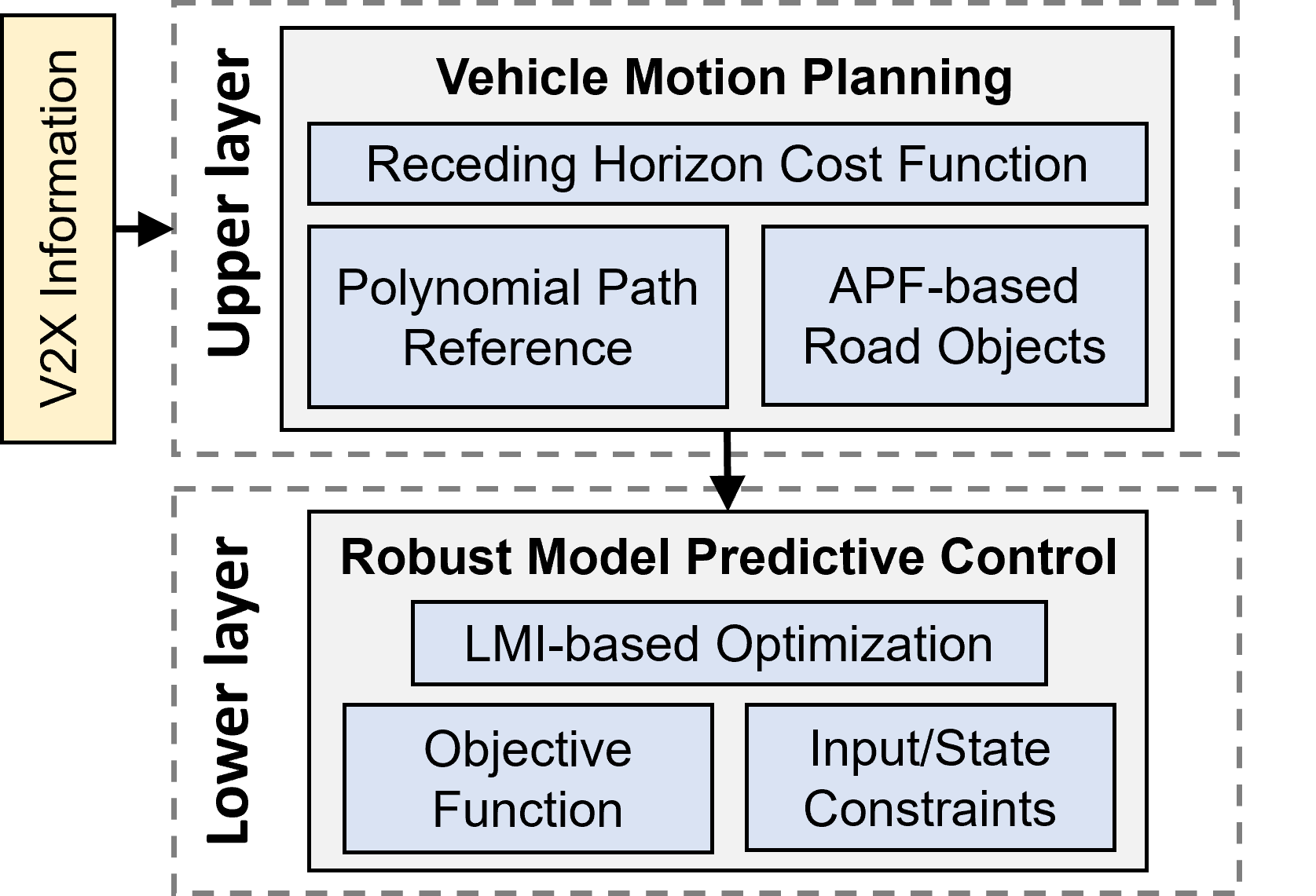}
    \caption{\label{fig.proposed.diagram} Hierarchical motion planning and control strategies.}
\end{figure}

\subsection{Architecture of Hierarchical Framework}
To address the proposed challenges in the aforementioned subsection, we introduce a hierarchical strategy of motion planning and offline RMPC approaches, as shown in Fig.~\ref{fig.proposed.diagram}. First, The upper layer holds a motion-planning function, which plays a vital role in collision avoidance. When receiving collision avoidance signals from V2X technology, the motion planning function first will generate a path via the high-order polynomial equation to avoid obstacles in normal scenarios. Moreover, in an emergency, V2X technology announces potentially dangerous signals. At that time, the fifth polynomial path will be modified to regenerate an optimal trajectory via the APF algorithm that captures any road objects in the artificial potentials. Finally, by satisfying the constraint's robustness, the lower layer plays a significant function in tracking a generated optimal trajectory. Offline-constrained RMPC is utilized by using LMI optimization with high tracking performance, high stability, and reduced computational burden. Therefore, the AV's driving is reliable and stable without any car crashes on the road.

\section{Motion Planning Approach}

\subsection{Objective Function}
The optimal trajectory will be generated by minimizing the cost function, including three penalties, including tracking reference, input, and interaction penalties, see, \cite{nguyen2023linear}. The objective function is formulated as
\begin{equation}
J_{\text{traj}} = \sum\limits_{k = 1}^{N_{\text{traj}}} {\left\| {y - {y_{\text{ref}}}} \right\|_{{Q}_{\text{traj}}}^2}  + \sum\limits_{k = 0}^{N_{\text{traj}} - 1} {\left\| {{u}_{\text{traj}}} \right\|_{{{R}_{\text{traj}}}}^2}  + {S}_{\text{traj}} J_{\text{syn}}\Comma
\end{equation}
\noindent where $y_{\text{ref}}$ denotes a fifth-degree polynomial path, see, \cite{liu2022interaction}, for the lateral position reference; ${u}_{\text{traj}}$ means the input signal of motion planning; $J_{\text{syn}}$ represents the potential values of the road object’s 3D map, i.e., road marks, road boundaries, and static/dynamic obstacles $(J_{\text{syn}} = J_{\text{obs}}^{\left( {{x_{\text{glo}}},{y_{\text{glo}}}} \right)} + J_{\text{lane}}^{\left( {{x_{\text{glo}}},{y_{\text{glo}}}} \right)} + J_{\text{road}}^{\left( {{x_{\text{glo}}},{y_{\text{glo}}}} \right)})$; ${Q}_{\text{traj}}$, ${R}_{\text{traj}}$, and ${S}_{\text{traj}}$ are adjustable weighting matrices, in which ${S}_{\text{traj}}$ is the most important emphasized the avoiding collision ability. Besides, the prediction horizon is set up equally with the control horizon (i.e., $N_{\text{traj}}$).

\subsection{Traffic-behavioral Obstacle Formulation}
The APF is utilized to capture any traffic behaviors, i.e., static/dynamic obstacles, road boundaries, and road marks, by different artificial potentials, see, \cite{nguyen2023linear}, formulated as
\begin{subequations}
    \begin{align}
        & J_{\text{obs}}^{\left( {{x_{\text{glo}}},{y_{\text{glo}}}} \right)} = \sum\limits_{j_o} {{A_{\text{obs}}}{e^{ - {{\left\{ {\frac{{{{\left( {{x_{\text{glo}}} - x_{\text{obs}}^{j_o}} \right)}^2}}}{{2\sigma _x^{j_o}}} + \frac{{{{\left( {{y_{\text{glo}}} - y_{\text{obs}}^{j_o}} \right)}^2}}}{{2\sigma _y^{j_o}}}} \right\}}^c}}}} , \\
        & J_{\text{lane}}^{\left( {{x_{\text{glo}}},{y_{\text{glo}}}} \right)} = \sum\limits_{k_l} {{A_{\text{lane}}}{e^{\left\{ {\frac{{ - {{\left( {{y_{\text{glo}}} - y_{\text{lane}}^{k_l}} \right)}^2}}}{{{d^2}}}} \right\}}}} , \\
        & J_{\text{road}}^{\left( {{x_{\text{glo}}},{y_{\text{glo}}}} \right)} = \frac{1}{2}\eta {\left\{ {\frac{1}{{{y_{\text{glo}}} - y_{\text{road}}^{\max }}} - \frac{1}{{{y_{\text{glo}}} - y_{\text{road}}^{\min }}}} \right\}^2},
    \end{align}
\end{subequations}
\noindent where $\left\{ x_{\text{glo}},y_{\text{glo}} \right\}$ represents the AV's global coordinate, which is set to be identical to the local coordinate of AV (i.e., $\left\{ x,y \right\}$); $A_{\text{obs}}$ and $A_{\text{lane}}$ are the tunable maximum obstacle and lane potential values; $\left\{ x_{\text{obs}}^{j_o},y_{\text{obs}}^{j_o} \right\}$ denotes the $j_o^{th}$ obstacle's location; $y_{\text{lane}}^{k_l}$ and $d$ are the $k_l^{th}$ lane road lateral coordinate and distance from the AV to the road mark; $c$ and $\eta$ reflect the adjustable coefficient of the obstacle shape and lane potential gain; ${y_{\text{road}}^{\min/\max }}$ represents the minimum/maximum of the road boundary, respectively. Additionally, $\sigma _x^{j_o}$ and $\sigma _y^{j_o}$ represent the object's longitudinal and lateral convergence coefficients, formulated in the following:
\begin{subequations}
\begin{align}
& {\sigma _x^{j_o} = \begin{cases*}
  \min _{\left\{ {{{( {\dot x - \dot x_{\text{obs}}^{j_o}})}^2},{{( {{x_{\text{glo}}} - x_{\text{obs}}^{j_o}})}^2}} \right\}}, \text{ if } {x_{\text{glo}}} \le x_{\text{obs}}^{j_o} \Comma\\
  \left( {L_{\text{obs}}^{j_o} + \varepsilon } \right)^2 , \begin{array}{*{20}{c}}
{}&{}&{}&{}&{}&{}
\end{array} \text{ otherwise}\Comma
\end{cases*}} \\
& {\sigma _y^{j_o} = {{\left( {\dfrac{{w_{\text{obs}}^{j_o}}}{2}} \right)}^2} ,}
\end{align}
\end{subequations}
\noindent where $\varepsilon$ is a safety factor guaranteeing the car from the obstacle's edges; $L_{\text{obs}}$ and $w_{\text{obs}}$ feature the obstacle's length and width, respectively.

\section{Offline Constrained RMPC Design}
Consider the formulation that minimizes the min-max cost function, see, \cite{park2011output}, at each time step $k$ as follows:
\begin{subequations} \label{eq.objective.func}
    \begin{align} 
    \label{eq.min_max}
    &\min_{{{\Delta} {u}(k+i)}} ~ \max_{{\left[ {\mathbf{A}(\rho (k)),\mathbf{B}(\rho (k))} \right] \in \Omega, k \geq 0}} J_{\infty} (k) \Comma
    \\ 
    &\text{subject to. } \left| {{{\Delta} {u}}\left( {k+i} \right)} \right| \le {{\Delta} {u}_{\max }} \Comma \label{eq.input.constraints} \\
    &\text{ } \text{ } \text{ } \text{ } \text{ } \text{ } \text{ } \text{ } \text{ } \text{ } \text{ } \text{ } \text{ } \left| {{\bm{\xi}}\left( {k+i} \right)} \right| \le {\bm{\xi}_{\max }} \Comma \label{eq.output.constraints}
    \end{align}
\end{subequations}
\noindent where ${J_\infty }\left( k \right) = \sum\limits_{i = 0}^\infty  {\left[ {\left\| {{\bm{\xi}} \left( {k+i} \right)} \right\|_{\mathbf{\bar Q}}^2 + \left\| {{\Delta} {u}\left( {k+i} \right)} \right\|_{{\bar R}}^2} \right]}$ with $i = 1, \dots, N$ is the horizon (i.e., $N = \infty$); besides, $\bm{\xi}$ denotes the state variables of the augmented model, while $\mathbf{\bar Q} \succ 0$ and ${\bar R} \succ 0$ denote the weighting matrices.

Let us define the Lyapunov function $V\left( {\bm{\xi}} \left( k+i \right)  \right)$ as
\begin{equation} \label{eq.quadratic.form}
    V\left( {\bm{\xi}} \left( k+i \right) \right) = \left\| {{\bm{\xi}} \left( k+i \right)} \right\|_{\mathbf{P}}^2 , \text{ } \mathbf{P} > 0 \Comma
\end{equation}
\noindent where $\mathbf{P}$ is a symmetric positive definite weighting matrix.

Suppose that $V\left( {\bm{\xi}} \left( k+i \right) \right)$ satisfies the Lyapunov condition, see, \cite{wan2003efficient}, with $\forall \left[ {\mathbf{A}\left( \rho(k) \right),\mathbf{B}\left( \rho(k) \right)} \right] \in \Omega ,k \ge 0$, which is described as follows:
\begin{equation} \label{eq.Lyapunov.sta}
    \begin{array}{l}
V\left( {{\bm{\xi}} \left( {k+i+1} \right)} \right) - V\left( {{\bm{\xi}} \left( {k+i} \right)} \right) \le \\
\begin{array}{*{20}{c}}
{}&{}&{}&{}
\end{array}{ - \left\| {{\bm{\xi}} \left( {k+i} \right)} \right\|_{\mathbf{\bar Q}}^2 - \left\| {{\Delta} {u} \left( {k+i} \right)} \right\|_{{\bar R}}^2} \FullStop
\end{array}
\end{equation}

The Lyapunov condition \eqref{eq.Lyapunov.sta} is imposed from zero to infinity (i.e., $k = 0: \infty $) to ensure the system's stability. Therefore, it will be required ${\bm{\xi}} (\infty + i) = 0$ or $V({\bm{\xi}} (\infty + i)) = 0$, obtained as $-V\left( {{\bm{\xi}} \left( {k+i} \right)} \right) \le -{J_\infty }\left( k \right)$.

Let us define a scalar $\gamma$ satisfying $V\left( {{\bm{\xi}} \left( {k+i} \right)} \right) \le \gamma$. Hence, we have the following:
\begin{equation} \label{eq.condition}
    {{\mathop {\max }\limits_{\left[ {\mathbf{A}\left( {\rho \left( {k} \right)} \right),\mathbf{B}\left( {\rho \left( {k} \right)} \right)} \right] \in \Omega ,k \ge 0}} {J_\infty }\left( k \right) \le V\left( {{\bm{\xi}} \left( {k+i} \right)} \right)} \le \gamma \Comma
\end{equation}

The delivered task now of the objective function \eqref{eq.min_max} aims to minimize $\gamma$ when satisfying the condition \eqref{eq.condition} as
\begin{subequations} \label{eq.new.cost.func}
    \begin{align} 
    &\min_{\gamma, \mathbf{P}} \gamma \Comma
    \\
    &\text{subject to. } \eqref{eq.input.constraints}, \eqref{eq.output.constraints}, \eqref{eq.Lyapunov.sta}, \text{ and } \eqref{eq.condition} \FullStop
    \end{align}
\end{subequations}

First, let $\mathbf{Q} = \gamma \mathbf{P}^{-1}$ and $\mathbf{Y} = \mathbf{K}\mathbf{Q}$, if the symmetric matrix $\mathbf{U}_{\text{cons}}$ exists $\mathbf{U}_{\text{cons}} = {{\Delta} {u}_{\max}^2\mathbf{I}}$, the input constraints \eqref{eq.input.constraints} can be expressed via the Euclidean norm as an LMI form, see, \cite{wan2003efficient}, as
\begin{equation} \label{eq.final.input.cons}
    \left[ {\begin{array}{*{20}{c}}
        {\mathbf{U}_{\text{cons}}}&{\mathbf{Y}^{{\bm{\top}}}} \\
        {\mathbf{Y}}&\mathbf{Q}
        \end{array}} \right] \ge 0 \FullStop
\end{equation}

Corresponding to \eqref{eq.final.input.cons}, the state constraints \eqref{eq.output.constraints} are also written in the following LMI form:
\begin{equation} \label{eq.final.output.cons}
    {\left[ {\begin{array}{*{20}{c}}
    {{{\mathbf{X}}_{{\text{cons}}}}}&{\left({\mathbf{A}}{\mathbf{Q}} + {\mathbf{B}}{\mathbf{Y}}\right)^{\top}}\\
    {{\mathbf{A}}{\mathbf{Q}} + {\mathbf{B}}{\mathbf{Y}}}&{\mathbf{Q}}
\end{array}} \right]} \ge 0 \Comma
\end{equation}
\noindent where $\mathbf{X}_{\text{cons}}$ denotes the symmetric matrix that is constructed by $\mathbf{X}_{\text{cons}} = {\bm{\xi}_{\max}^2\mathbf{I}}$, see, \cite{wan2003efficient}.

In order to ensure uncertainties ${\Omega}$, the state constraints \eqref{eq.final.output.cons} consider uncertain matrices $\mathbf{A}\left( \rho(k) \right)$ and $\mathbf{B}\left( \rho(k) \right)$ with $\forall \left[ \mathbf{A}\left( \rho(k) \right), \mathbf{B}\left( \rho(k) \right) \right] \in \Omega$, expressed in the following:
\begin{equation} \label{eq.final.state.cons}
    {\left[ {\begin{array}{*{20}{c}}
    {{{\mathbf{X}}_{{\text{cons}}}}}&*\\
    {{\mathbf{A}}\left( j \right){\mathbf{Q}} + {\mathbf{B}}\left( j \right){\mathbf{Y}}}&{\mathbf{Q}}
\end{array}} \right]_{j = 1,\dots,4.}} \ge 0 \Comma
\end{equation}
\noindent where $*$ denotes the corresponding symmetric component.

Additionally, the Lyapunov function $V\left( {\bm{\xi}} \left( k+i \right) \right)$ is considered to calculate the control gain $\mathbf{K}$ when satisfying the Lyapunov stability condition \eqref{eq.Lyapunov.sta}. Therefore, by substituting the robust feedback control ${\Delta} {u}$ in the augmented system \eqref{eq.augmented.system}, the Lyapunov stability condition \eqref{eq.Lyapunov.sta} can be rewritten as
\begin{equation} \label{eq.robust.stability}
    \left\| {{\bm{\xi}} \left( k+i \right)} \right\|_{\left( {\scriptstyle\left\| {\mathbf{A} + \mathbf{B} \mathbf{K}} \right\|_{\mathbf{P}}^2 - \mathbf{P} + \mathbf{\bar Q} + \left\| \mathbf{K} \right\|_{{\bar R}}^2\hfill} \right)}^2 \le 0 \FullStop
\end{equation}

Inequality equation \eqref{eq.robust.stability} can be calculated equivalently as
\begin{equation} \label{eq.short.robust.stability}
    \begin{array}{l}
    \left\| {\mathbf{A} + \mathbf{B} \mathbf{K}} \right\|_{\mathbf{P}}^2 - \mathbf{P} + \mathbf{\bar Q} + \left\| \mathbf{K} \right\|_{{\bar R}}^2 \le 0 \FullStop
    \end{array}
\end{equation}

Furthermore, we consider the LPV model system \eqref{eq.dis.veh.model} with $\forall \left[ {\mathbf{A}\left( \rho(k) \right),\mathbf{B}\left( \rho(k) \right)} \right] \in \Omega$, substituting $\mathbf{Q} = \gamma \mathbf{P}^{ - 1}$ and $\mathbf{Y} = \mathbf{K}\mathbf{Q}$ into the stability condition \eqref{eq.short.robust.stability}, which is satisfied at each vertex in the following symmetric matrix:
\begin{equation}
    {\left[ {\begin{array}{*{20}{c}}
    {\mathbf{Q}}&*&*&*\\
    {{\mathbf{A}}\left( j \right){\mathbf{Q}} + {\mathbf{B}}\left( j \right){\mathbf{Y}}}&{\mathbf{Q}}&*&*\\
    {{\mathbf{\bar Q}^{1/2}}{\mathbf{Q}}}&{\mathbf{0}}&{\gamma {\mathbf{I}}}&*\\
    {{{\bar R}^{1/2}}{\mathbf{Y}}}&{\mathbf{0}}&{\mathbf{0}}&{\gamma {\mathbf{I}}}
    \end{array}} \right]_{j = 1,\dots,4.}} \ge 0 \FullStop
\end{equation}

Finally, based on the LMI optimization, the inequality condition \eqref{eq.condition} is rewritten equivalently as
\begin{equation}
    \left[ {\begin{array}{*{20}{c}}
    1&{{\bm{\xi}} {{\left( {k+i} \right)}^{{\bm{\top}}}}}\\
    {{\bm{\xi}} \left( {k+i} \right)}&\mathbf{Q}
    \end{array}} \right] \ge 0, \text{ } \mathbf{Q} > 0 \FullStop
\end{equation}

Now, an efficient offline-constrained RMPC is derived by using the asymptotically stable invariant ellipsoid, see, \cite{wan2003efficient}, when considering the discrete-time system (${\bm{\xi}} (k+1)$).

The uncertain discrete-time LPV system \eqref{eq.dis.veh.model} is subject to input and state constraints, i.e., \eqref{eq.final.input.cons} and \eqref{eq.final.state.cons}. After that, giving the initial state ${\bm{\xi}}(0)$, and following:

$\textbf{\textit{Step 1}}$: Compute minimizers $\gamma$, $\mathbf{Q}(k)$, ${\bm{\xi}}(k)$, and $\mathbf{Y}(k)$, by using the objective function \eqref{eq.new.cost.func} with an additional condition $\mathbf{Q}(k - 1) > \mathbf{Q}(k)$, store $\mathbf{Q}(k)$ and $\mathbf{Y}(k)$ in the look-up table.

$\textbf{\textit{Step 2}}$: If $k < N$, choose a state ${\bm{\xi}}(k+1)$ satisfying $\left\| {{\bm{\xi}}(k + 1)} \right\|_{\mathbf{Q}^{-1}}^2 \le 1$. Then, let us put $k:=k+1$ and turn back $\textbf{\textit{Step 1}}$.

We can obtain the robust control gain via the look-up table technique, i.e., $\mathbf{K} = \mathbf{Y} \mathbf{Q}^{-1}$, see proof in \cite{wan2003efficient}.
Eventually, the control signal of the augmented model can be calculated by ${\Delta} {u}(k) = \mathbf{K}{\bm{\xi}}(k)$.

\section{Case Studies}
\begin{figure*}[t!]
    \centering
    \includegraphics[width = 170mm]{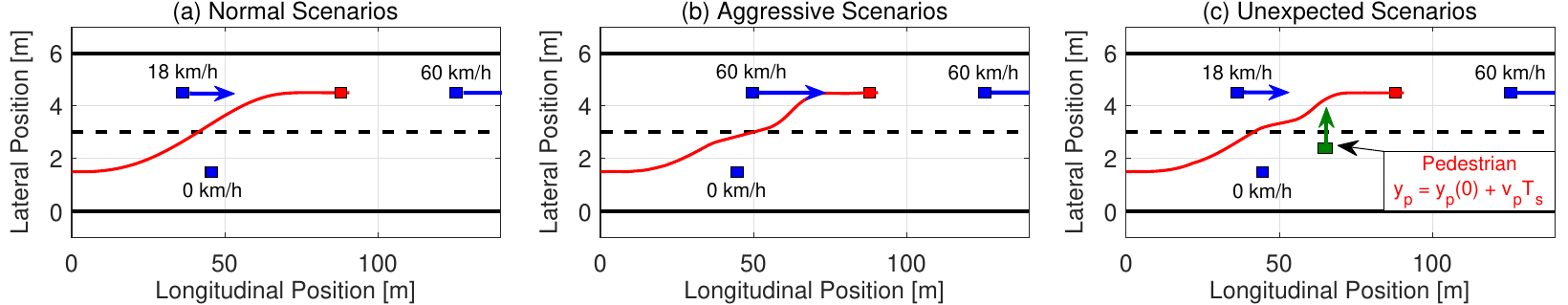}
    \caption{\label{fig.case.studies} Case studies: (a) Normal scenarios, (b) Aggressive scenarios, and (c) Unexpected scenarios with a simple pedestrian speed profile.}
\end{figure*}

In this section, various scenarios are considered to emphasize the efficiency of our proposed approach in handling harsh road conditions. Fig.~\ref{fig.case.studies} depicts three case studies, i.e., normal, aggressive, and unexpected scenarios.


\subsection{Baseline Controllers}
To emphasize the superiority of our proposed approach, three alternative robust MPCs are introduced as follows:

\subsubsection{Online Constrained Robust Model Predictive Control \textbf{(Online)}, see, \cite{wan2003efficient}}
The LPV system \eqref{eq.dis.veh.model} is considered subject to input and state constraints \eqref{eq.input.constraints} and \eqref{eq.output.constraints} at each time $k$. Therefore, by minimizing the objective function \eqref{eq.new.cost.func}, we obtain the control gain $\mathbf{K}(k+i)$.

\subsubsection{Offset Offline Constrained Robust Model Predictive Control \textbf{(Offset offline)}}
A steady-state approach can be utilized to improve the tracking performance at each sampling time. Therefore, the improved robust feedback control is expressed as ${\Delta} {u}(k+i) = \mathbf{K} {\bm{\xi}}(k+i) + {\Delta} {u}^{o}(k)$, which has emphasized by the steady-state control (i.e., ${\Delta} {u}^{o}(k)$), which is calculated by solving the following steady-state condition as ${\bm{\xi}}^{o}(k+1) = \mathbf{A} {\bm{\xi}}^{o}(k) + \mathbf{B} {\Delta} {u}^{o}(k) + \mathbf{E} {{\dot \psi} _{\text{ref}}},$ where ${\bm{\xi}}^{o}$ denotes the nominal state variables.

\begin{figure}[t!]
    \centering
    \includegraphics[width = 90mm]{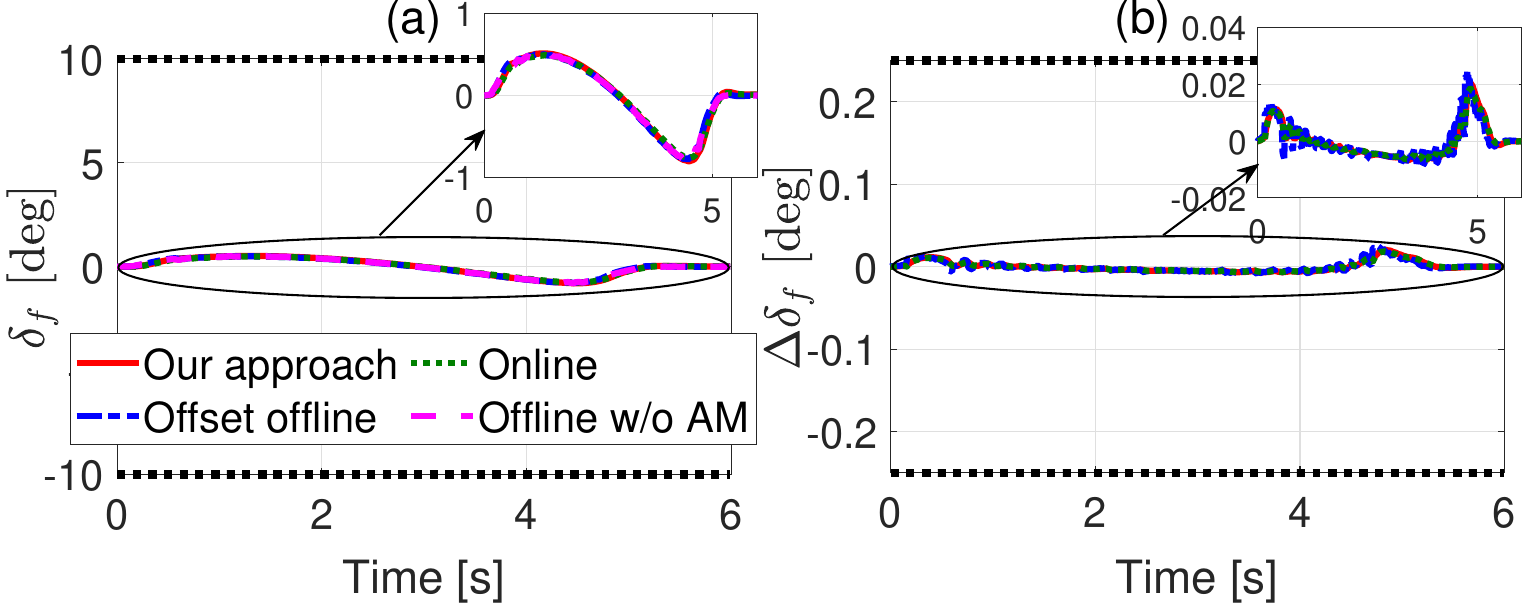}
    \caption{\label{fig.case.1.steer} Input parameters: (a) Steering wheel angle and (b) Steering wheel angle rate in normal scenarios.}
\end{figure}

\begin{figure}[t!]
    \centering
    \includegraphics[width = 95mm]{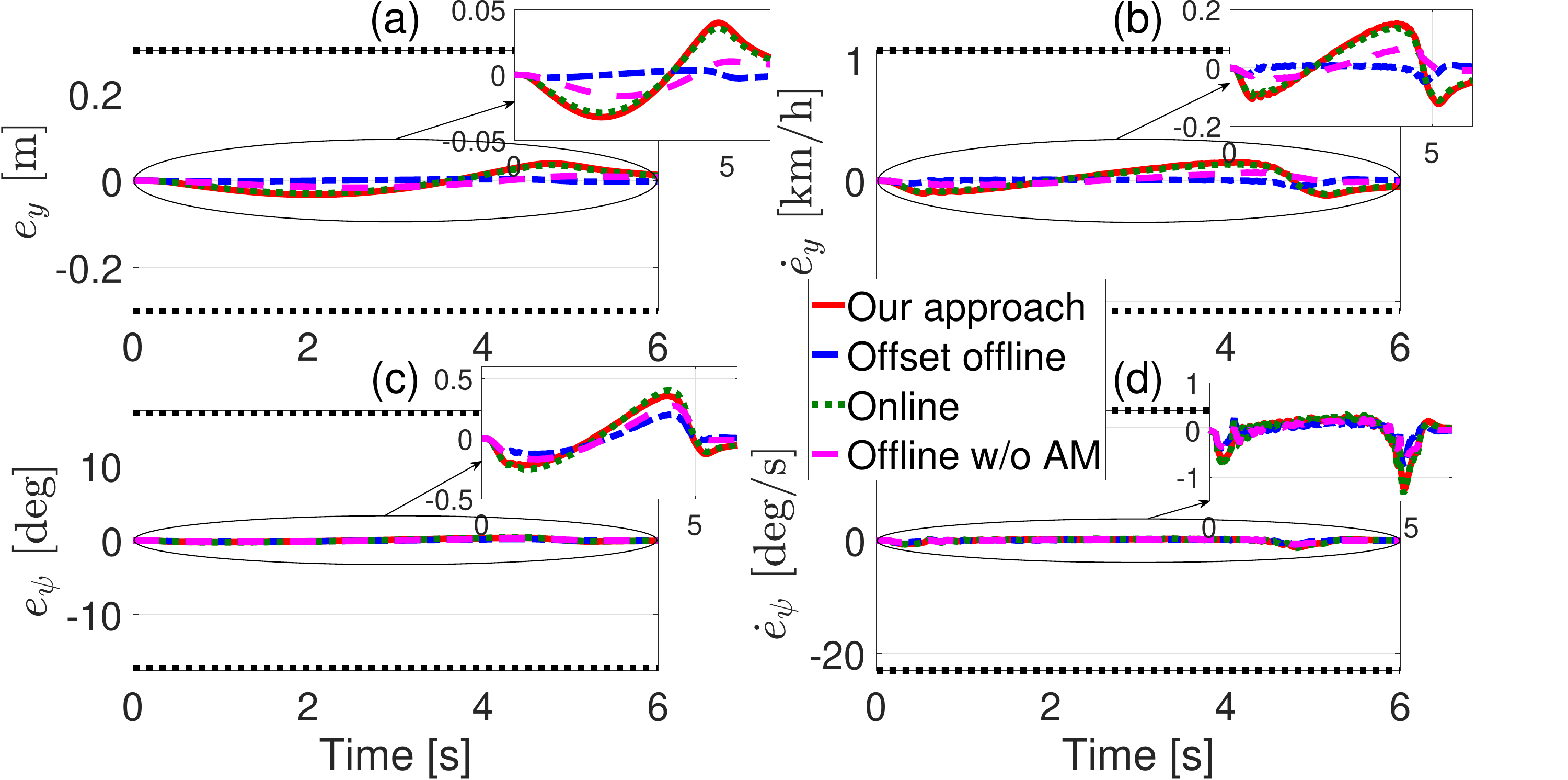}
    \caption{\label{fig.case.1.state} State parameters: (a) Lateral position error, (b) Lateral velocity error, (c) Yaw angle error, and (d) Yaw rate error in normal scenarios.}
\end{figure}

\begin{table}[!t]
\centering
\caption{Time execution of all methods.} \label{fig.burden}
\setlength{\tabcolsep}{2pt}
\resizebox{\columnwidth}{!}{
\begin{tabular}{c c c c c}
\hline
Method          & Our approach & Offset-offline & Offline w/o AM & Online  \\\hline \hline
Average time [ms]    & 1.16 & 1.37 & 1.14 & 1.84$\times 10^3$  \\\hline
Maximum time [ms]    & 3.26 & 2.51 & 1.57 & 2.53$\times 10^3$  \\\hline
\end{tabular}}
\end{table}

\subsubsection{Offline Constrained Robust Model Predictive Control Without Augmented Model \textbf{(Offline w/o AM)}, see, \cite{nam2023robust}}
Instead of using the extended model \eqref{eq.ext.discrete.veh.model}, the offline RMPC method is employed via the tracking vehicle model \eqref{eq.discrete.veh.model}. Therefore, the system input is the steering wheel angle (i.e., ${u} = {\delta}_f$), which is defined as ${u}(k) = \mathbf{K}{\bm{\xi}}(k)$.

\subsection{Simulation Results}
After receiving the potential risk signal from V2X technology, AV performs a lane-changing action to avoid collisions when driving under various road adhesion coefficients in complex scenarios at a relatively high speed ($v_x = 15$m/s).

\begin{figure}[t!]
    \centering
    \includegraphics[width = 75mm]{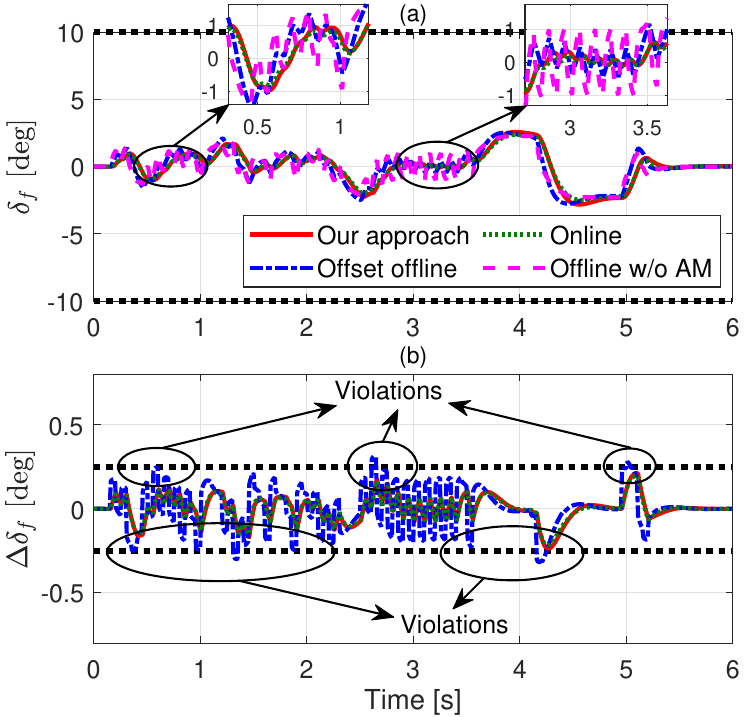}
    \caption{\label{fig.case.2.steer} Input parameters: (a) Steering wheel angle and (b) Steering wheel angle rate in aggressive scenarios.}
\end{figure}

\begin{figure}[t!]
    \centering
    \includegraphics[width = 95mm]{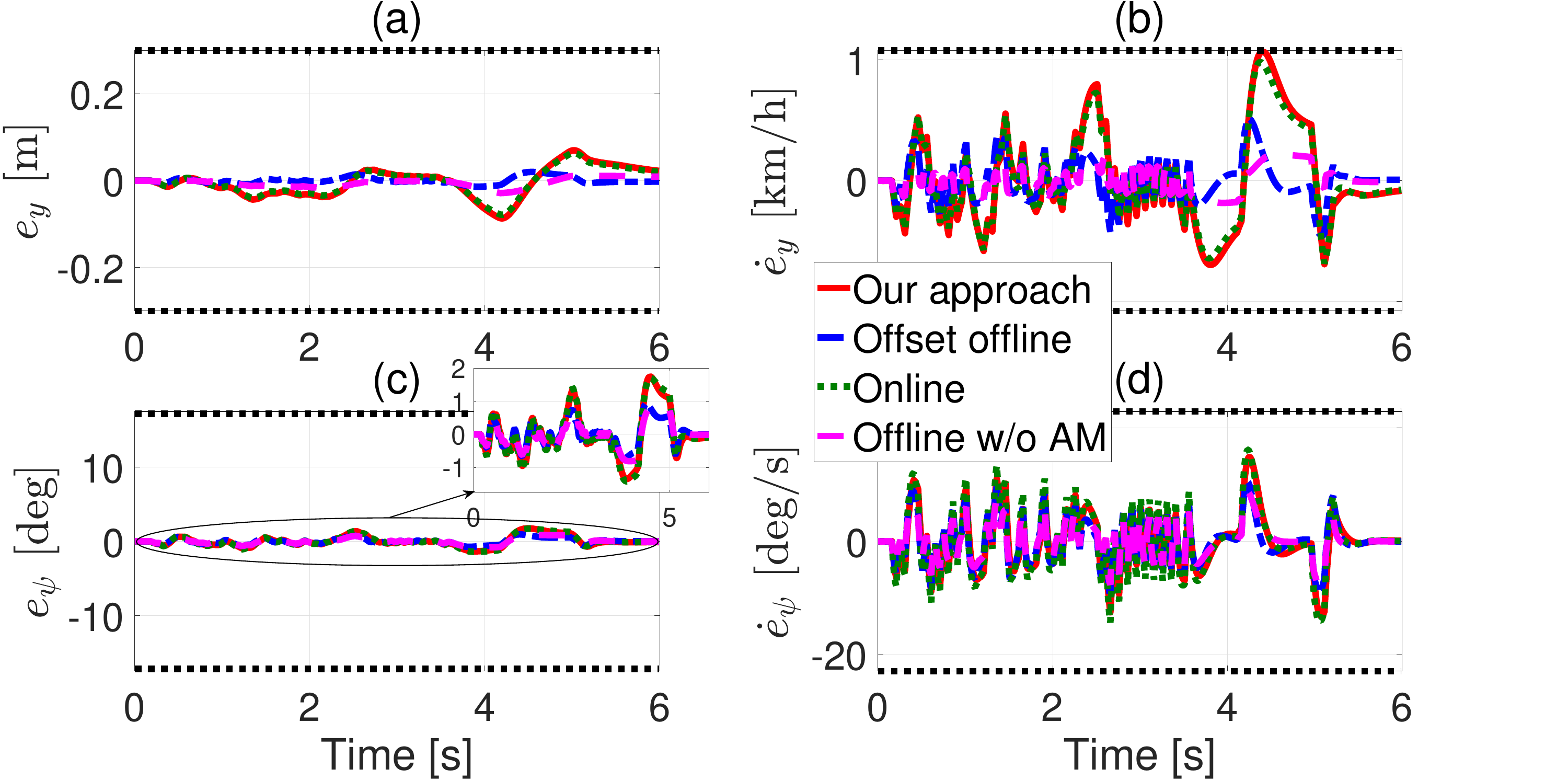}
    \caption{\label{fig.case.2.state} State parameters: (a) Lateral position error, (b) Lateral velocity error, (c) Yaw angle error, and (d) Yaw rate error in aggressive scenarios.}
\end{figure}

In normal scenarios, by satisfying the input and state constraints, shown Figs.~\ref{fig.case.1.steer} and ~\ref{fig.case.1.state}, our proposed approach has achieved a high efficiency correspondingly compared with the results of the online RMPC method while improving the time execution, depicted in Tab.~\ref{fig.burden}. However, as illustrated in Figs.~\ref{fig.case.1.state}(a), (b), (c), and (d), outstanding features of the steady-state algorithm and the vehicle tracking model are emphasized, the tracking performances of the offset-offline RMPC approach and offline RMPC approach w/o AM are expressed significantly more than our proposed and online RMPC methods.

\begin{figure}[t!]
    \centering
    \includegraphics[width = 75mm]{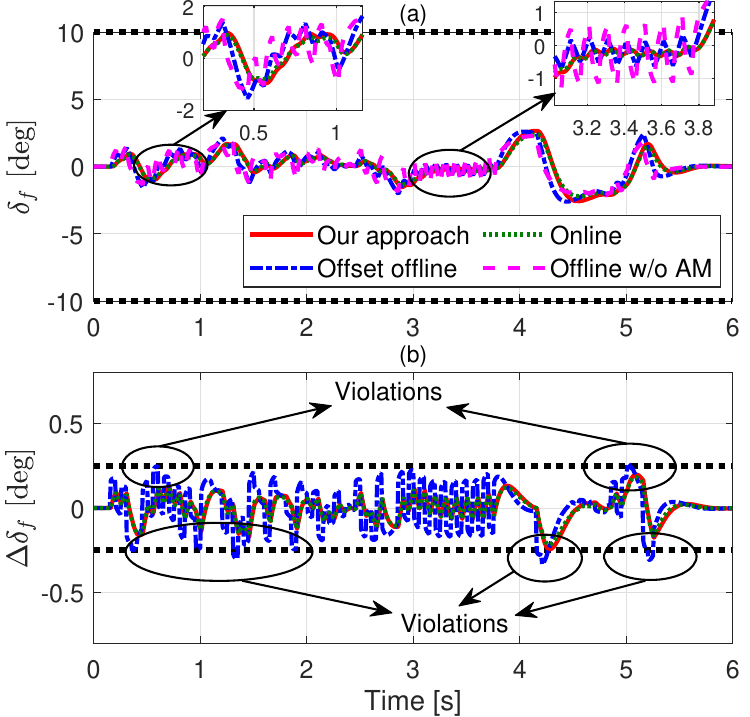}
    \caption{\label{fig.case.3.steer} Input parameters: (a) Steering wheel angle and (b) Steering wheel angle rate in unexpected scenarios.}
\end{figure}

\begin{figure}[t!]
    \centering
    \includegraphics[width = 95mm]{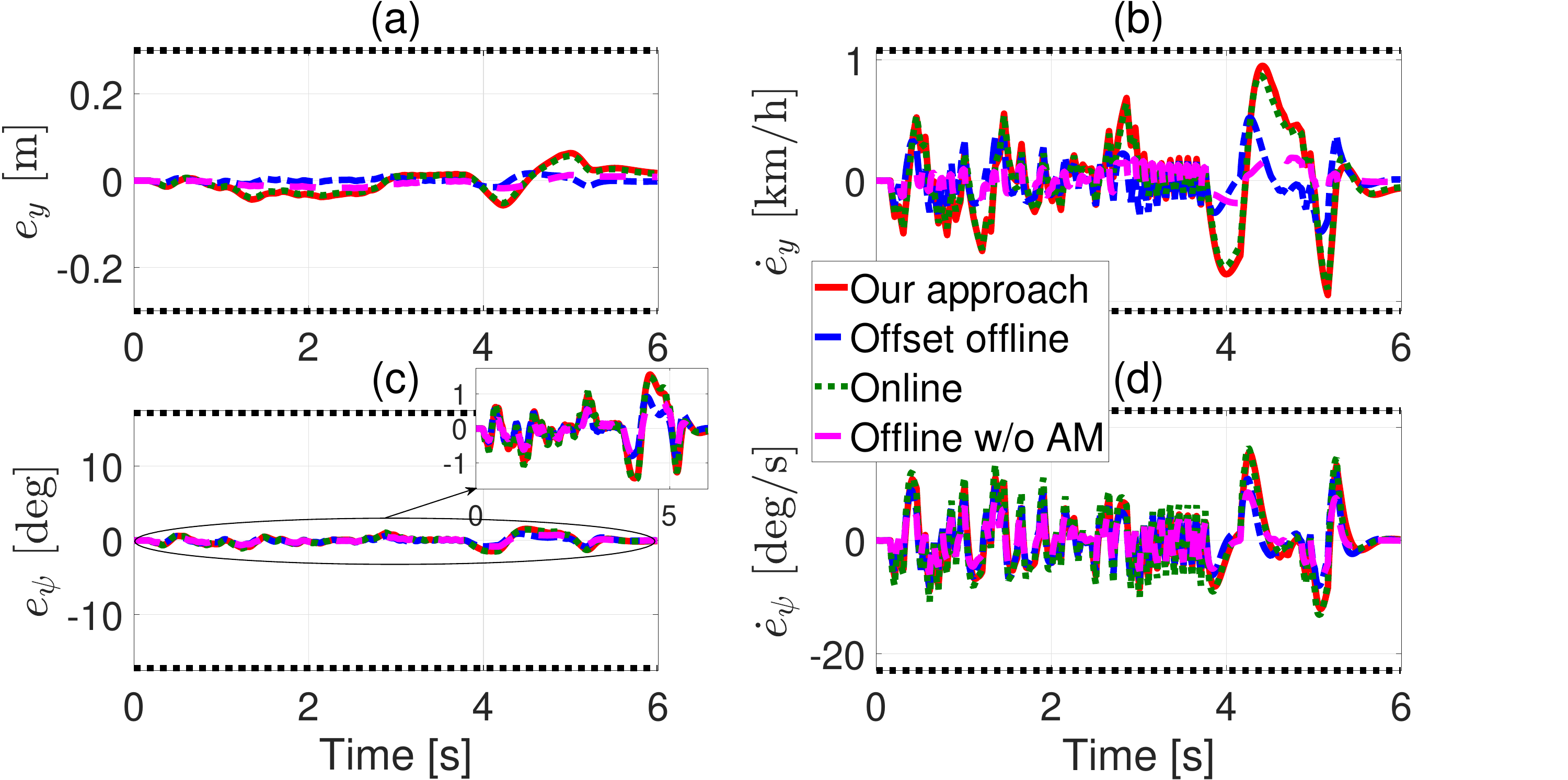}
    \caption{\label{fig.case.3.state} State parameters: (a) Lateral position error, (b) Lateral velocity error, (c) Yaw angle error, and (d) Yaw rate error in unexpected scenarios.}
\end{figure}

Although the offset-offline RMPC approach and offline RMPC approach w/o AM achieve high performance in normal situations, our aim focuses on complex and sudden situations when driving, see Figs.~\ref{fig.case.studies}(b) and (c). Therefore, our proposed and online RMPC methods have emphasized the reasonable handles in both tracking performance and improving input vibrations by satisfying the input and state constraints, shown in Figs.~\ref{fig.case.2.steer}, \ref{fig.case.2.state}, \ref{fig.case.3.steer}, and \ref{fig.case.3.state}. In contrast, the steady-state approach and offline RMPC method w/o AM are prioritized in the tracking efficiency, which leads to input violations and input vibrations significantly, as depicted in Fig.~\ref{fig.case.2.steer}(b) and \ref{fig.case.3.steer}(b), thereby the steering wheel angles are unrealistic in real-time. Therefore, in these cases, our proposed method achieved an outstanding ability to balance performances of tracking performance as well as the computational burden, illustrated in Tab.~\ref{fig.burden}, when driving in complex scenarios compared with three existing RMPC methods (i.e., offset-offline, online, and offline w/o AM).

\section{Conclusion}
This study proposed a hierarchical strategy for AVs when considering uncertain parameters and driving in complex scenarios. By using IDM, HV's behaviors are modeled as the car-following model, then observed and perceived from V2X technology. Whenever receiving potentially dangerous signals, the upper layer determines the environment and captures road objects comprehensively via the APF method, so an optimal trajectory will be generated to avoid collisions. After generating an optimal trajectory, in the lower layer, an offline-constrained RMPC is employed to track this optimal trajectory, besides, by satisfying the input and state constraints robustly the AV achieved high performance in tracking and stability when compared with three existing RMPCs (i.e., offset-offline, online, and offline w/o AM).

\bibliographystyle{IEEEtran}

\bibliography{reference.bib}

\end{document}